# Fault Diagnosis for Underwater Vehicles using Moving Horizon Estimation and Gaussian Processes

Fotis Panetsos and Kostas J. Kyriakopoulos

*Abstract*— **This work proposes a model-based fault detection and diagnosis framework for underwater vehicles subject to actuator faults that explicitly accounts for the presence of unmodeled dynamics. To this end, a Moving Horizon Estimator (MHE) is developed to estimate the lumped disturbance, capturing both unmodeled and fault effects. Gaussian Processes (GPs) are employed to approximate the unmodeled dynamics, providing predictions of the corresponding mean and uncertainty across diverse operating conditions. During online operation, the residual between the MHE lumped disturbance estimate and the GP prediction is evaluated using a Generalized Likelihood Ratio Test. By incorporating GP-based predictions within the diagnostic framework, robustness to unmodeled dynamics is achieved, enabling effective fault detection and isolation as well as accurate quantitative estimation of fault magnitude. The proposed methodology is experimentally validated in a laboratory water tank, demonstrating reliable diagnostic performance under both open-loop and closed-loop control.**

## I. INTRODUCTION

The field of underwater robotics has gained increasing attention in recent years due to the versatility of underwater vehicles (UVs), which enables their deployment in a wide range of applications, including offshore oil and gas operations [1], aquaculture [2], marine ecosystem monitoring [3], and inspection of submerged infrastructures [4]. However, the complexity of these missions, combined with the inherent challenges of underwater environments, increases the likelihood of faults and compromises the safety of UVs. Such faults may degrade system performance, lead to premature mission termination, or even result in vehicle loss [5].

Among the various subsystems, thrusters are particularly prone to failure since they are directly exposed to the surrounding environment. Mechanical wear, biofouling, corrosion, and debris entanglement may cause partial or complete loss of thrust, thereby hindering mission execution. Consequently, the development of reliable fault detection and diagnosis (FDD) techniques for UVs is essential to enable timely fault identification and enhance operational safety.

### A. Related Works

The problem of FDD for actuator faults in UVs has been extensively studied in the literature. Generally, FDD aims to detect the occurrence of a fault, isolate the faulty component, and estimate the fault severity. Based on the underlying methodology, existing approaches can be classified into model-free (data-driven) and model-based techniques [6].

The authors are with the Center for AI & Robotics (CAIR) and the Electrical Eng. Program, Engineering Division, New York University Abu Dhabi, Abu Dhabi, United Arab Emirates. E-mails: {f.panetsos, kkyria}@nyu.edu

Data-driven FDD approaches leverage learning-based methods to detect and characterize faults directly from measured data without relying on an explicit vehicle model. In [7], a deep learning-based FDD framework employing a Sequence Convolutional Neural Network (SeqCNN) is proposed and trained using measurements collected from the “Haizhe” UV under both normal and faulty operating conditions. Experimental results demonstrate reliable classification of different fault types, such as slight and severe propeller damage. Using the same benchmark dataset, subsequent works have explored alternative learning architectures to further improve diagnostic performance. In [8], a hybrid framework integrating an improved Parrot Optimizer with a Random Forest classifier is introduced, achieving enhanced diagnostic accuracy. More recently, [9] proposes a multisensor feature fusion Kolmogorov–Arnold Network (MFKAN) that demonstrates superior performance compared to previous methods on the same “Haizhe” dataset.

However, despite their promising performance, the aforementioned model-free approaches rely on fault data for training, which are often difficult to obtain in practice. In addition, being formulated as classification frameworks, these methods identify fault types without providing a quantitative estimate of the fault magnitude. Finally, their performance may degrade under operating conditions that differ from the training data distribution, thereby limiting their generalization capability and reliability in safety-critical missions.

In contrast to data-driven techniques, model-based approaches exploit the vehicle dynamics for fault diagnosis. In [10], the known dynamic model governing the vehicle horizontal motion is used to construct a bank of extended Kalman filters (EKFs), each corresponding to a distinct operating hypothesis, i.e., normal conditions or complete failure of a thruster. Residuals generated by the EKFs are evaluated against predefined thresholds, enabling fault identification, as demonstrated through real-world experiments. In [11], a switching-mode hidden Markov model is employed to represent normal vehicle behavior and multiple failure modes, with thruster anomalies characterized as insufficient or zero thrust. A particle filter operating on this model, along with fixed mode transition probabilities, provides robust state estimation, as validated in sea-trial experiments.

Instead of relying on multiple models or modes, disturbance estimation techniques can be adopted for FDD, since the reduced control input caused by the unknown thrust loss can be interpreted as an equivalent disturbance acting on the system dynamics. In this context, [12] augments the vehicle state vector with the fault-induced disturbance and employs a Gaussian particle filter for joint estimation. Fault detection is performed by processing the estimated reduced input using a Bayesian algorithm and comparing the resulting quantity with a predefined threshold. In [13], an integral extended state observer is proposed to estimate the total system uncertainty, including fault effects, and this estimate is subsequently integrated into a fault-tolerant control scheme for trajectory tracking. Moreover, [14] employs a high-order sliding mode observer to estimate the lumped disturbance, including fault-related terms. Under severe fault conditions, conditionally triggered thruster control allocation and fault estimation are activated to compensate for faults and ensure robust control performance.

However, the performance of model-based FDD techniques strongly depends on the accuracy of the assumed dynamics. In practice, parameter uncertainties, simplifying assumptions, and hydrodynamic effects that cannot be fully captured from first principles introduce discrepancies between the nominal and the true system behavior. In addition, these unmodeled dynamics vary across operating conditions, further hindering the ability to distinguish fault-induced disturbances from unmodeled effects. As a result, the selection of appropriate detection thresholds becomes challenging, and ambiguities may arise during fault isolation. Furthermore, the presence of unmodeled dynamics degrades the accuracy of fault magnitude estimation, particularly in the case of partial thruster failure.

Consequently, approximation of the unmodeled dynamics is essential for robust fault diagnosis and estimation of fault severity. In this context, Gaussian Processes (GPs) have emerged as an effective nonparametric probabilistic framework for learning modeling errors while simultaneously quantifying predictive uncertainty [15], [16], [17]. These properties make GPs particularly well suited for model-based FDD, where distinguishing fault-induced disturbances from unmodeled dynamics is critical.

### B. Contributions

In this work, we design a model-based FDD framework for UVs capable of reliably detecting actuator faults, isolating the faulty thruster, and accurately estimating the corresponding loss of effectiveness despite the presence of unmodeled dynamics. Specifically, a Moving Horizon Estimator (MHE) is formulated based on the nominal UV model to estimate the total disturbance acting on the system, encompassing both unmodeled dynamics and fault-induced effects. Under fault-free conditions, Gaussian Processes (GPs) are trained to learn the unmodeled dynamics, providing a nonparametric probabilistic representation across varying operating regimes. Using the MHE estimates and GP predictions, a residual is generated with statistical properties determined by the GP predictive uncertainty, thereby enabling adaptation to changing operating conditions. A Generalized Likelihood Ratio Test (GLRT) is subsequently formulated for each thruster to process this residual, detect faults, isolate the faulty actuator under the single-fault assumption, and estimate fault severity via maximum likelihood estimation. The probabilistic approximation of the unmodeled dynamics enables straightforward threshold selection, rapid fault detection, reduced false alarms, and accurate fault magnitude estimation. The effectiveness of the proposed methodology is experimentally validated in a laboratory water tank using the VideoRay Defender vehicle.

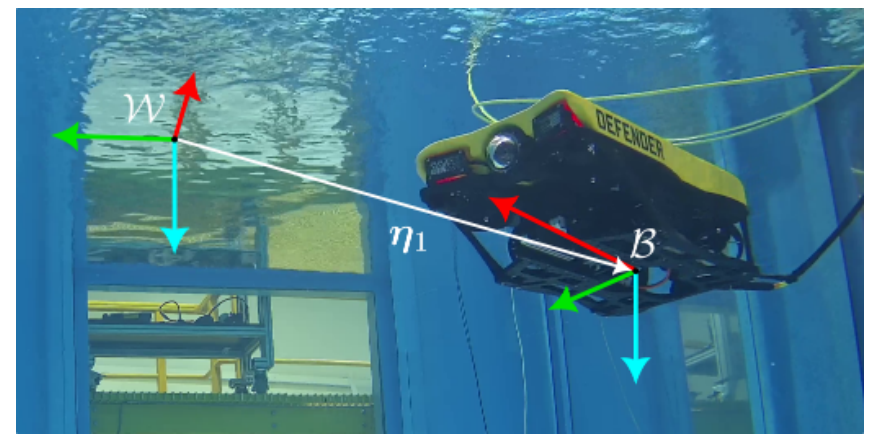

Fig. 1. The VideoRay Defender vehicle inside the water tank. The world frame $\mathcal{W}$, the body-fixed frame $\mathcal{B}$, and the vehicle position $\boldsymbol{\eta}_1$ are depicted, with the $x$, $y$, and $z$ axes of each frame indicated by red, green, and blue arrows, respectively.

## II. Preliminaries

### A. Equations of Motion

This section presents the dynamic model governing the motion of the UV, as described in [18].

Consider the VideoRay Defender vehicle employed in this study, and let $\mathcal{W}$ and $\mathcal{B}$ denote the world frame and the body-fixed frame, respectively, as shown in Fig. 1. The origin of $\mathcal{W}$ is defined at a fixed point on the water surface of the test tank, while the body-fixed frame $\mathcal{B}$ is attached to the vehicle. The pose of the UV is represented by $\boldsymbol{\eta} = \begin{bmatrix}\boldsymbol{\eta}_1^\top & \boldsymbol{\eta}_2^\top\end{bmatrix}^\top$, where $\boldsymbol{\eta}_1 = \begin{bmatrix}x & y & z\end{bmatrix}^\top$ denotes the position and $\boldsymbol{\eta}_2 = \begin{bmatrix}\phi & \theta & \psi\end{bmatrix}^\top$ denotes the orientation expressed using the Euler roll–pitch–yaw convention. The kinematics of the UV are given by:

$$\dot{\boldsymbol{\eta}} = \boldsymbol{J}(\boldsymbol{\eta})\boldsymbol{v}, \tag{1}$$

where $\boldsymbol{v} = \begin{bmatrix}\boldsymbol{v}_1^\top & \boldsymbol{v}_2^\top\end{bmatrix}^\top$ is the vehicle velocity w.r.t. $\mathcal{B}$, with $\boldsymbol{v}_1 = \begin{bmatrix}u & v & w\end{bmatrix}^\top$ representing the linear velocities in surge, sway, and heave, and $\boldsymbol{v}_2 = \begin{bmatrix}p & q & r\end{bmatrix}^\top$ representing the roll, pitch, and yaw rates. The matrix $\boldsymbol{J}(\boldsymbol{\eta})$ defines the transformation from body-fixed velocities in $\mathcal{B}$ to $\mathcal{W}$.

The dynamics of the UV are governed by:

$$\boldsymbol{M}\dot{\boldsymbol{v}} + \boldsymbol{C}\left(\boldsymbol{v}\right)\boldsymbol{v} + \boldsymbol{D}\boldsymbol{v} + \boldsymbol{g}\left(\boldsymbol{\eta}\right) = \boldsymbol{\tau} + \boldsymbol{d}_n, \tag{2}$$

where $\boldsymbol{M} \in \mathbb{R}^{6\times6}$ is the inertia matrix including both rigid-body and added-mass effects, $\boldsymbol{C}\left(\boldsymbol{v}\right) \in \mathbb{R}^{6\times6}$ denotes the Coriolis and centripetal matrix, $\boldsymbol{D} \in \mathbb{R}^{6\times6}$ is the hydrodynamic damping matrix, $\boldsymbol{g}\left(\boldsymbol{\eta}\right) \in \mathbb{R}^6$ is the restoring force and moment vector due to gravity and buoyancy, $\boldsymbol{d}_n \in \mathbb{R}^6$ represents unmodeled dynamics, and $\boldsymbol{\tau} \in \mathbb{R}^6$ denotes the control forces and moments applied to the vehicle.

### B. Actuation Model

The VideoRay Defender vehicle is actuated by seven thrusters, comprising four horizontal thrusters that control the horizontal motion of the UV, namely the surge, sway, and yaw DoFs, and three vertical thrusters responsible for the heave, roll, and pitch motions. Let $\boldsymbol{l}_j \in \mathbb{R}^3$ and $\boldsymbol{e}_j \in \mathbb{R}^3$ denote the position vector and thrust direction of the $j$-th thruster w.r.t. $\mathcal{B}$, respectively. The generalized control input vector $\boldsymbol{\tau}$ applied to the vehicle is then computed as:

$$\boldsymbol{\tau} = \underbrace{\begin{bmatrix} \boldsymbol{e}_1 & \cdots & \boldsymbol{e}_m \\ \boldsymbol{l}_1 \times \boldsymbol{e}_1 & \cdots & \boldsymbol{l}_m \times \boldsymbol{e}_m \end{bmatrix}}_{\boldsymbol{B}} \boldsymbol{u}, \tag{3}$$

where $m$ is the total number of thrusters, $\boldsymbol{u} = \begin{bmatrix} u_1 & \cdots & u_m \end{bmatrix}^\top \in \mathbb{R}^m$ is the vector of thruster forces and $\boldsymbol{B} \in \mathbb{R}^{6\times m}$ is the thruster allocation matrix. It should be noted that the transient dynamics of the thrusters are assumed to be negligible with respect to the overall vehicle dynamics.

### C. Thruster Fault Modeling

Our goal is to detect and isolate potential faults that affect the vehicle thrusters and lead to a loss of effectiveness. In particular, thruster effectiveness is modeled by the diagonal matrix $\boldsymbol{\Lambda} = \text{diag}(\lambda_1, \ldots, \lambda_m) \in \mathbb{R}^{m\times m}$, where $\lambda_j \in [0, 1]$ denotes the health state of the $j$-th thruster, with $\lambda_j = 1$ corresponding to a fully operational thruster and $\lambda_j = 0$ indicating complete failure. Consequently, in the presence of faults, the actual control input applied to the vehicle is given by:

$$\boldsymbol{\tau} = \boldsymbol{B}\boldsymbol{\Lambda}\boldsymbol{u}. \tag{4}$$

Considering that the health state of the thrusters is unknown, the UV dynamics in Eq. 2 are reformulated as:

$$\boldsymbol{M}\dot{\boldsymbol{v}} + \boldsymbol{C}(\boldsymbol{v})\boldsymbol{v} + \boldsymbol{D}\boldsymbol{v} + \boldsymbol{g}(\boldsymbol{\eta}) = \boldsymbol{B}\boldsymbol{u} + \boldsymbol{d}, \tag{5}$$

where $\boldsymbol{d} \in \mathbb{R}^6$ is the total lumped disturbance vector capturing both unmodeled dynamics and fault-induced effects, i.e., $\boldsymbol{d} = \boldsymbol{d}_n + \boldsymbol{d}_f$, with $\boldsymbol{d}_f = \boldsymbol{B}(\boldsymbol{\Lambda} - \boldsymbol{I}_m)\boldsymbol{u}$ denoting the fault-induced disturbance and $\boldsymbol{I}_m \in \mathbb{R}^{m\times m}$ the identity matrix. It should be highlighted that the term $\boldsymbol{d}_n$ accounts for unmodeled dynamics and parameter uncertainties, while external environmental disturbances are assumed to be negligible.

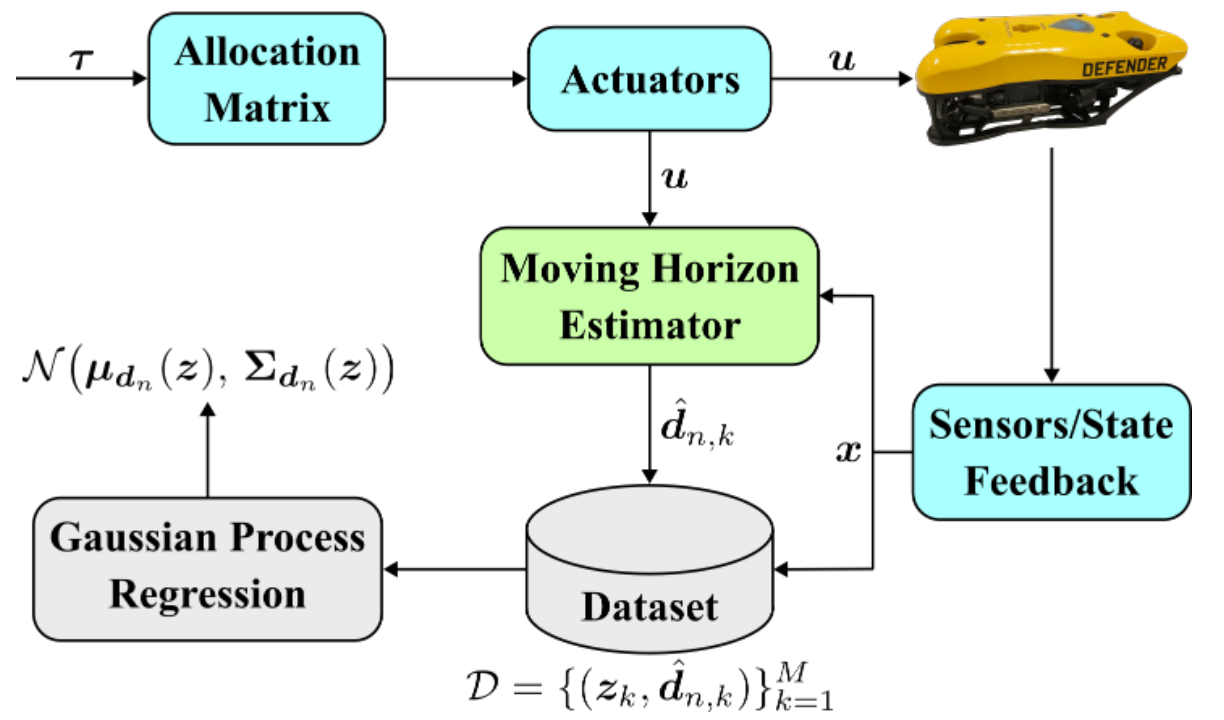

Fig. 2. Data collection under normal operation for approximating the unmodeled dynamics using GPs.

## III. Problem Statement

The objective of this work is to develop a model-based FDD framework that enables reliable identification of a single actuator fault, remains robust to unmodeled dynamics and parameter uncertainties, and provides quantitative estimation of the corresponding fault magnitude $\lambda_j$. To this end, the proposed methodology integrates model-based estimation, data-driven learning, and statistical hypothesis testing into a unified framework composed of the following components:

*1) Moving Horizon Estimator:* An MHE is formulated to obtain a dynamics-consistent estimate of the lumped disturbance vector $\boldsymbol{d}$, aggregating the contributions of unmodeled dynamics $\boldsymbol{d}_n$ and potential fault-induced effects $\boldsymbol{d}_f$.

*2) Gaussian Processes:* GP models are employed to learn the unmodeled dynamics $\boldsymbol{d}_n$ from MHE disturbance estimates collected under fault-free operation ($\boldsymbol{d}_f = \boldsymbol{0}$), as illustrated in Fig. 2. The resulting GP representation provides a probabilistic characterization of the expected behavior of $\boldsymbol{d}_n$ across diverse operating conditions.

*3) Generalized Likelihood Ratio Test:* By comparing the lumped disturbance estimated by the MHE with the probabilistic GP prediction of the unmodeled dynamics, a residual signal is constructed whose statistical properties are governed by the state-dependent GP predictive uncertainty. A GLRT is then formulated for each thruster to assess whether the residual is consistent with fault-free operation or indicative of an actuator fault, thereby enabling fault detection and isolation under the assumption of at most one fault. Finally, the GLRT formulation provides a maximum likelihood estimate (MLE) of the corresponding fault magnitude $\lambda_j$, enabling quantitative evaluation of actuator degradation.

An overview of the proposed framework is shown in Fig. 3.

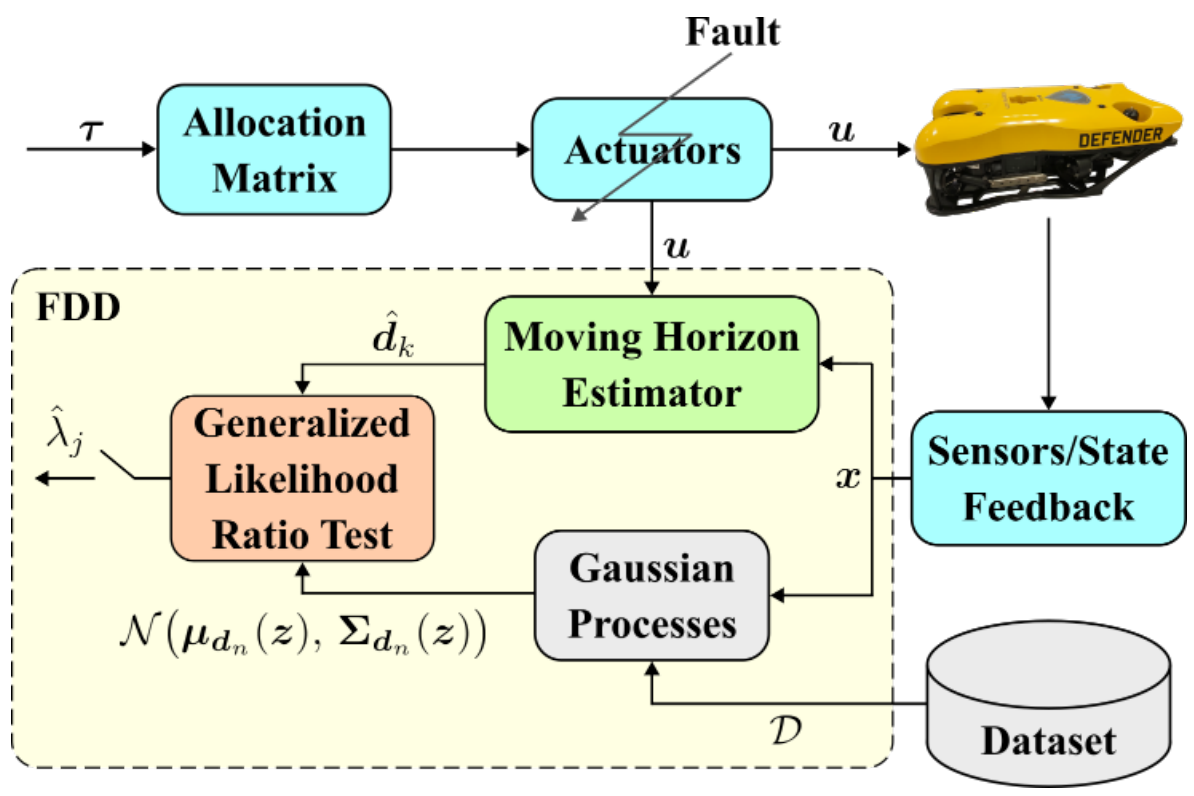

Fig. 3. The proposed fault detection and diagnosis framework.

## IV. Methodology

### A. Moving Horizon Estimator

Under both normal and faulty operating conditions, an MHE is employed in order to estimate the lumped disturbance $\boldsymbol{d}$. To this end, the disturbance dynamics are modeled as a random walk according to:

$$\dot{\boldsymbol{d}} = \boldsymbol{w}, \tag{6}$$

where $\boldsymbol{w} \in \mathbb{R}^6$ denotes the process noise. This formulation provides a simple yet effective means of capturing unknown and time-varying disturbances [19].

By augmenting the UV state with the disturbance vector $\boldsymbol{d}$, i.e., $\boldsymbol{x} = \begin{bmatrix} \boldsymbol{\eta}^\top & \boldsymbol{v}^\top & \boldsymbol{d}^\top \end{bmatrix}^\top$, the augmented system dynamics are defined based on Eqs. 1, 5, and 6 as:

$$\begin{aligned} \dot{\boldsymbol{x}} &= \boldsymbol{f}\left(\boldsymbol{x}, \boldsymbol{u}, \boldsymbol{w}\right), \\ \boldsymbol{y} &= \boldsymbol{h}\left(\boldsymbol{x}\right) + \boldsymbol{\omega}, \end{aligned} \tag{7}$$

where $\boldsymbol{y}$ denotes the measurement vector consisting of the UV pose $\boldsymbol{\eta}$ and velocity $\boldsymbol{v}$, $\boldsymbol{h}(\cdot)$ is the measurement function, and $\boldsymbol{\omega}$ denotes the measurement noise.

The MHE is an optimization-based technique that estimates the system state by solving a constrained optimization problem over a finite receding horizon. At each sampling instant $t \geq N$, where $N$ denotes the horizon length, the MHE optimizes over $N+1$ states $\boldsymbol{x}_{t-N:t} = \{\boldsymbol{x}_k\}_{k=t-N}^{t}$ and $N$ process noise terms $\boldsymbol{w}_{t-N:t-1} = \{\boldsymbol{w}_k\}_{k=t-N}^{t-1}$, given $N+1$ measurements $\{\boldsymbol{y}_k\}_{k=t-N}^{t}$ and $N$ control inputs $\{\boldsymbol{u}_k\}_{k=t-N}^{t-1}$. The optimal estimates at time $t$, denoted by $\hat{\boldsymbol{x}}_{t-N:t|t}$ and $\hat{\boldsymbol{w}}_{t-N:t-1|t}$, are obtained by solving the following optimization problem:

$$\begin{aligned} \min_{\boldsymbol{x}_{t-N:t},\, \boldsymbol{w}_{t-N:t-1}} \quad & \frac{1}{2} \left\| \boldsymbol{x}_{t-N} - \hat{\boldsymbol{x}}_{t-N|t-1} \right\|_{\boldsymbol{P}}^2 \\ & + \frac{1}{2} \sum_{k=t-N}^{t} \left\| \boldsymbol{y}_k - \boldsymbol{h}(\boldsymbol{x}_k) \right\|_{\boldsymbol{R}}^2 \\ & + \frac{1}{2} \sum_{k=t-N}^{t-1} \left\| \boldsymbol{w}_k \right\|_{\boldsymbol{Q}}^2 \\ \text{s.t.} \quad & \boldsymbol{x}_{k+1} = \boldsymbol{\phi}(\boldsymbol{x}_k, \boldsymbol{u}_k, \boldsymbol{w}_k, \mathrm{d}t), \\ & k = t-N, \dots, t-1, \end{aligned} \tag{8}$$

where $\mathrm{d}t$ is the sampling period, $\boldsymbol{\phi}(\cdot)$ denotes the discrete-time equivalent model of $\boldsymbol{f}(\cdot)$ (Eq. 7), $\boldsymbol{P}$, $\boldsymbol{R}$, and $\boldsymbol{Q}$ are positive definite weighting matrices associated with the arrival cost, measurement residual, and process noise, respectively, and $\hat{\boldsymbol{x}}_{t-N|t-1}$ denotes the prior state estimate at the beginning of the moving horizon, computed at time $t-1$. For the sake of brevity, let $\hat{\boldsymbol{d}}_k = \hat{\boldsymbol{d}}_{k|k}$ denote the MHE estimate of the lumped disturbance at sampling instant $k$.

### B. Gaussian Process Regression

The estimation of the lumped disturbance is not sufficient for reliable FDD due to the presence of unmodeled dynamics; therefore, GPs are employed to learn $\boldsymbol{d}_n$.

In general, GPs provide a nonparametric Bayesian framework for approximating an unknown scalar function $g(\boldsymbol{z}) : \mathbb{R}^{n_z} \to \mathbb{R}$ [20]. Consider a set of $M$ noisy observations of $g$ at input locations $\boldsymbol{z}_k \in \mathbb{R}^{n_z}$, given by:

$$y_k = g(\boldsymbol{z}_k) + \upsilon_k, \qquad \upsilon_k \sim \mathcal{N}(0, \sigma_n^2), \tag{9}$$

for $k = 1, \dots, M$. The corresponding training dataset is defined as $\mathcal{D} = \{(\boldsymbol{z}_k, y_k)\}_{k=1}^{M}$. By specifying a GP prior with kernel $k(\cdot,\cdot)$ and conditioning on the dataset $\mathcal{D}$, the posterior distribution of the function value at a query input $\boldsymbol{z}$ is also Gaussian, i.e.,

$$g(\boldsymbol{z}) \mid \mathcal{D} \sim \mathcal{N}\big(\mu(\boldsymbol{z}),\, \sigma^2(\boldsymbol{z})\big), \tag{10}$$

where $\mu(\boldsymbol{z})$ and $\sigma^2(\boldsymbol{z})$ denote the predictive mean and variance, respectively.

However, standard GP regression is associated with high computational cost, since the evaluation of the posterior variance scales quadratically, $\mathcal{O}(M^2)$, with the number of data points in $\mathcal{D}$. Considering that accurately capturing diverse dynamic regimes entails a sufficiently large dataset, real-time deployment of GPs becomes infeasible. In addition, since GPs approximate only scalar-valued functions, independent GP models must be employed for each component of $\boldsymbol{d}_n$, further increasing the computational burden and necessitating the use of an approximation technique.

Consequently, stochastic variational Gaussian Process (SVGP) models are employed. In particular, the SVGP framework introduces inducing variables and leverages stochastic variational inference to obtain a scalable approximation, enabling the application of GPs to large datasets while maintaining adequate predictive accuracy, as detailed in [21], [22]. In contrast to standard GP regression, the computational complexity of SVGP depends solely on the number of inducing variables $\tilde{M}$ rather than the size of the full dataset, with $\tilde{M} \ll M$, and reduces the prediction cost to $\mathcal{O}(\tilde{M}^2)$, thereby rendering SVGP models suitable for real-time deployment.

To learn the unmodeled dynamics, data are collected under various operating conditions and in the absence of faults ($\boldsymbol{d}_f = \boldsymbol{0}$), such that the MHE disturbance estimates correspond to measurements of the unmodeled dynamics $\boldsymbol{d}_n$, according to Fig. 2. After training independent SVGP models for each component and stacking the resulting posterior distributions, the unmodeled dynamics are approximated as:

$$\boldsymbol{d}_n(\boldsymbol{z}) \mid \mathcal{D} \sim \mathcal{N}\big(\boldsymbol{\mu}_{\boldsymbol{d}_n}(\boldsymbol{z}),\, \boldsymbol{\Sigma}_{\boldsymbol{d}_n}(\boldsymbol{z})\big), \tag{11}$$

where $\boldsymbol{\mu}_{\boldsymbol{d}_n}(\boldsymbol{z})$ and $\boldsymbol{\Sigma}_{\boldsymbol{d}_n}(\boldsymbol{z})$ denote the predicted mean vector and diagonal covariance matrix, respectively, and the input $\boldsymbol{z}$ is constructed from the UV state, as outlined in Section V-A.

### C. Generalized Likelihood Ratio Test

Given the GP predictive mean of the unmodeled dynamics and the lumped disturbance estimate $\hat{\boldsymbol{d}}_k$ provided by the MHE at the sampling instant $k$, the following residual is defined:

$$\boldsymbol{r}_k = \hat{\boldsymbol{d}}_k - \boldsymbol{\mu}_{\boldsymbol{d}_n}(\boldsymbol{z}_k). \tag{12}$$

By leveraging the GP predictive covariance, the residual under normal operating conditions, i.e., $\lambda_j = 1$ for all $j = 1, \dots, m$, follows a zero-mean Gaussian distribution:

$$\mathcal{H}_0 : \quad \boldsymbol{r}_k \sim \mathcal{N}\big(\boldsymbol{0},\, \boldsymbol{\Sigma}_{\boldsymbol{d}_n}(\boldsymbol{z}_k)\big). \tag{13}$$

In contrast, in the presence of a single fault in the $j$-th thruster, i.e., $\lambda_j \in [0, 1)$, the residual exhibits a nonzero mean corresponding to the fault-induced disturbance $\boldsymbol{d}_f$. In this case, the residual distribution can be written as:

$$\mathcal{H}_1^{(j)} : \quad \boldsymbol{r}_k \sim \mathcal{N}\big(\Delta\lambda_j \boldsymbol{b}_j u_{j,k},\, \boldsymbol{\Sigma}_{\boldsymbol{d}_n}(\boldsymbol{z}_k)\big), \tag{14}$$

where $\Delta\lambda_j = \lambda_j - 1$, $\boldsymbol{b}_j$ is the $j$-th column of the thruster allocation matrix $\boldsymbol{B}$, and $u_{j,k}$ denotes the control input for the $j$-th thruster at sampling instant $k$.

The objective is to determine whether the UV operates under normal conditions ($\mathcal{H}_0$) or in the presence of a single fault in the $j$-th thruster ($\mathcal{H}_1^{(j)}$). To this end, the decision problem is addressed using the GLRT framework, a widely used statistical approach for composite hypothesis testing [23]. To enhance the robustness of GLRT-based detection against outliers, a moving window of length $W$ is considered by stacking the residual observations as $\tilde{\boldsymbol{r}} = \begin{bmatrix}\boldsymbol{r}_{k-W+1}^\top & \cdots & \boldsymbol{r}_k^\top\end{bmatrix}^\top$. Accordingly, for each thruster $j$, the hypothesis test can be expressed as:

$$\begin{aligned} \mathcal{H}_0: &\quad \tilde{\boldsymbol{r}} \sim \mathcal{N}(\mathbf{0}, \boldsymbol{\Sigma}), \\ \mathcal{H}_1^{(j)}: &\quad \tilde{\boldsymbol{r}} \sim \mathcal{N}(\boldsymbol{h}_j \Delta\lambda_j, \boldsymbol{\Sigma}), \end{aligned} \tag{15}$$

where $\boldsymbol{\Sigma} = \text{blkdiag}\big(\boldsymbol{\Sigma}_{\boldsymbol{d}_n}(\boldsymbol{z}_{k-W+1}), \ldots, \boldsymbol{\Sigma}_{\boldsymbol{d}_n}(\boldsymbol{z}_k)\big)$ is the covariance matrix of the stacked residual vector over the moving window and $\boldsymbol{h}_j = \begin{bmatrix}\boldsymbol{b}_j^\top u_{j,k-W+1} & \cdots & \boldsymbol{b}_j^\top u_{j,k}\end{bmatrix}^\top$.

The likelihood ratio comparing the likelihood of the observed data under the two hypotheses is defined as:

$$L_j(\tilde{\boldsymbol{r}}) = \frac{\sup_{\Delta\lambda_j} p\left(\tilde{\boldsymbol{r}} \mid \mathcal{H}_1^{(j)}, \Delta\lambda_j\right)}{p(\tilde{\boldsymbol{r}} \mid \mathcal{H}_0)}, \tag{16}$$

where $p(\cdot)$ denotes the Gaussian probability density function. In the GLRT, the unknown parameter $\Delta\lambda_j$ in $\mathcal{H}_1^{(j)}$ is replaced by its MLE, given by:

$$\begin{aligned} \hat{\Delta\lambda}_j &= \left(\boldsymbol{h}_j^\top \boldsymbol{\Sigma}^{-1} \boldsymbol{h}_j\right)^{-1} \boldsymbol{h}_j^\top \boldsymbol{\Sigma}^{-1} \tilde{\boldsymbol{r}} \\ \hat{\Delta\lambda}_j &= \max\left(-1, \min\left(\hat{\Delta\lambda}_j, 0\right)\right), \end{aligned} \tag{17}$$

considering that $\lambda_j \in [0, 1]$. Finally, by taking the logarithm of Eq. 16 and substituting the MLE (Eq. 17), the GLRT decides $\mathcal{H}_1^{(j)}$ if:

$$T_j(\tilde{\boldsymbol{r}}) = 2\hat{\Delta\lambda}_j \boldsymbol{h}_j^\top \boldsymbol{\Sigma}^{-1} \tilde{\boldsymbol{r}} - \hat{\Delta\lambda}_j^2 \boldsymbol{h}_j^\top \boldsymbol{\Sigma}^{-1} \boldsymbol{h}_j > \delta, \tag{18}$$

where $\delta$ is the test threshold for the GLRT statistic $T_j$.

The proposed approach enables both fault detection and isolation based on Eq. 18, as well as estimation of the fault magnitude using the MLE ($\hat{\lambda}_j = 1 + \hat{\Delta\lambda}_j$). Moreover, the GP predictive covariances provide adaptability by reflecting local uncertainties associated with the operating conditions, in contrast to conventional GLRT approaches that often assume a constant, manually tuned covariance matrix. Finally, it should be noted that, due to similarities among the columns of $\boldsymbol{B}$, the GLRT statistics corresponding to multiple thrusters may exceed the threshold even when only a single fault is present. In such cases, under the assumption of a single fault, isolation is achieved by selecting the thruster with the maximum value of $T_j$.

## V. Experimental Results

### A. Experimental Setup & Implementation Details

The experiments presented below were carried out in a laboratory water tank of $6.3\,\text{m} \times 3.7\,\text{m} \times 1.8\,\text{m}$, as shown in Fig. 4. Vehicle localization is achieved using an ArUco board [24] placed at the bottom of the tank and composed of markers of two different sizes to ensure reliable detection across multiple depths. The VideoRay Defender is equipped with a downward-looking camera for board detection, a depth sensor, and an AHRS that provides orientation and angular velocity measurements. The respective measurements are fused using a Kalman filter, and therefore an estimate of the UV state, namely pose and velocity, is available at each sampling instant. The vehicle thrusters are actuated by a Pulse Width Modulation (PWM) signal. To relate this normalized command to the generated thruster force $u_j$, a mapping was identified from experimental data. The resulting model consists of two polynomial expressions describing forward and reverse rotation, while accounting for a narrow dead-band region around zero input. Finally, communication with the vehicle sensors and thrusters is achieved through the tether. The software is implemented in the Robot Operating System (ROS) [25] and runs on a computer with an NVIDIA RTX 4090 GPU and 128 GB of RAM.

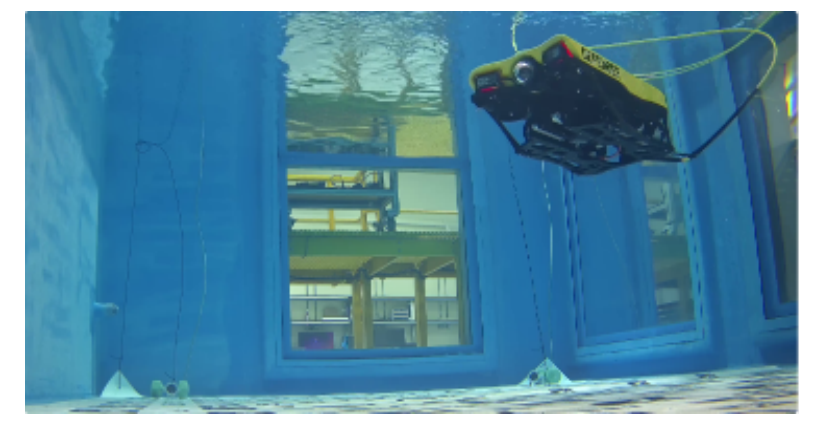

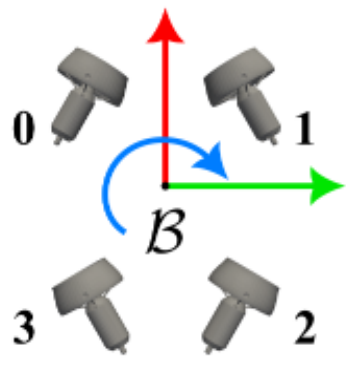

Fig. 4. The laboratory water tank and the configuration of the horizontal thrusters.

Throughout the experiments, faults are assumed to affect only the horizontal thrusters of the vehicle, whose configuration is illustrated in Fig. 4. For this purpose, a PID controller is used to regulate the vertical thrusters, maintaining constant depth and near-zero roll and pitch angles. As a result, a reduced-order dynamic model is adopted that captures the surge, sway, and yaw DoFs. It should be noted that, although horizontal motion is overactuated from a control perspective, fault isolation in the horizontal plane is more challenging than in the vertical subsystem, which is actuated by three thrusters. This increased difficulty stems from the similar contributions of the horizontal thrusters in the allocation matrix, resulting in partially overlapping fault signatures.

Regarding the implementation details and computational cost of the proposed framework, the dynamic parameters of the vehicle were identified from experimental data using a least-squares approach. The resulting nominal model is employed in the MHE formulated in CasADi [26]. Solving the optimization problem of Eq. 8 with a horizon length of $N = 15$ and a sampling rate of 25 Hz requires 7.8 ms on the aforementioned computer setup. As for the SVGP models, the GPyTorch library [27] is used to learn the unmodeled dynamics in the surge, sway, and yaw DoFs. Specifically, the GP input vector is selected as $\boldsymbol{z}_t = \{\begin{bmatrix}u_k & v_k & r_k & \dot{u}_k & \dot{v}_k & \dot{r}_k\end{bmatrix}\}_{k=t-N}^{t}$, ensuring consistency with the finite-horizon nature of the MHE disturbance estimates. Accelerations are incorporated into the GP input to account for parameter uncertainties in the inertia matrix and are obtained via cubic spline interpolation of the velocity measurements followed by differentiation of the fitted splines. The GP models are trained using a Matérn kernel, normalization, $\tilde{M} = 3000$ inducing points, and a dataset comprising more than $M = 100$k samples. Hyperparameters are learned by maximizing the Variational Evidence Lower Bound (ELBO). The mean execution time required to evaluate the posterior distribution of the three independent GP models is 20.2 ms. Finally, a moving window of 3 s, corresponding to $W = 75$ samples at 25 Hz, is adopted for the GLRT. The test threshold is set to $\delta = 150$ based on residual statistics obtained under normal operating conditions.

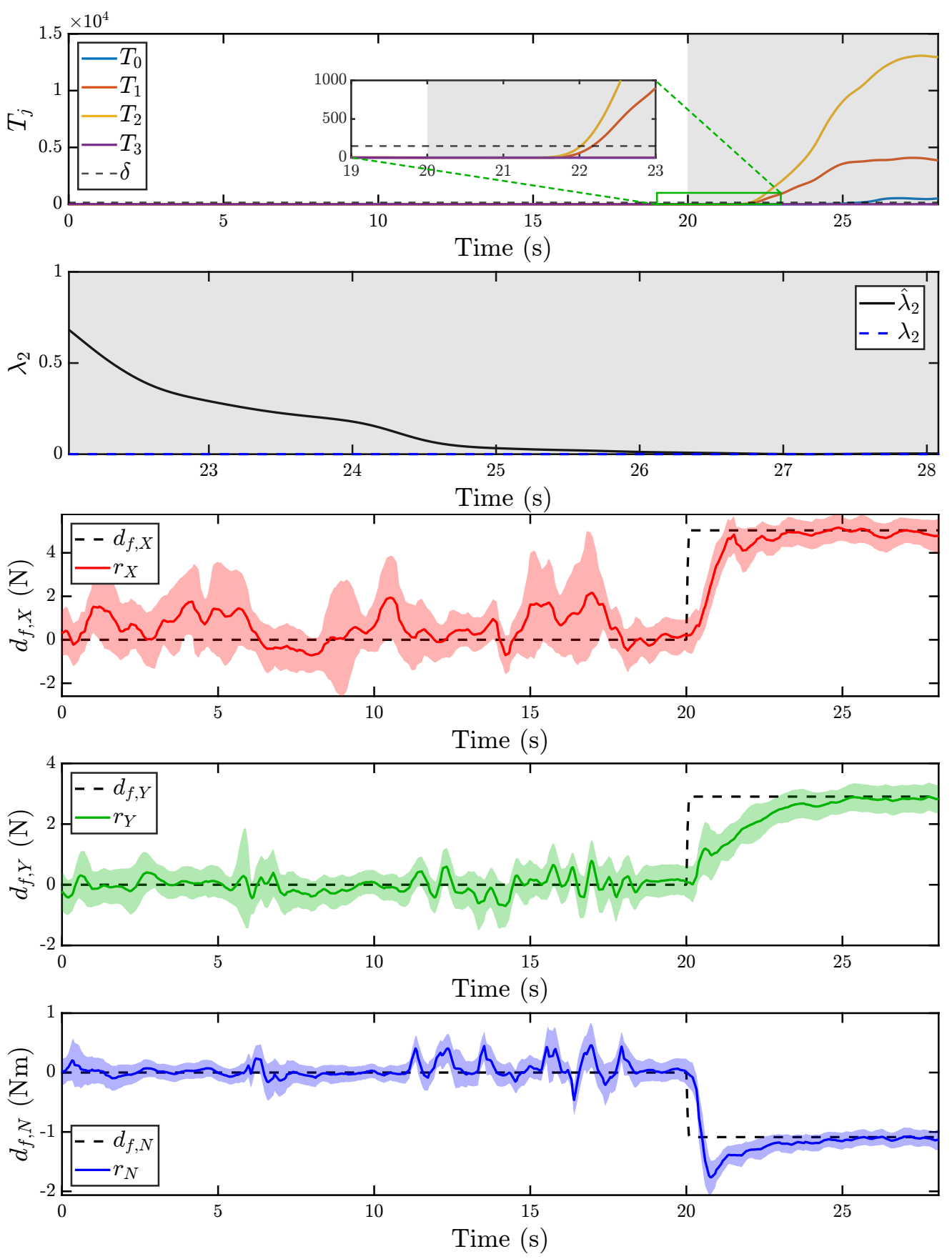

Fig. 5. First teleoperation experiment with complete failure of Thruster 2. From top to bottom: GLRT statistics $T_j$ of the horizontal thrusters, with the shaded region indicating operation after fault injection; the MLE estimate $\hat{\lambda}_2$ together with the true value after fault detection; and the residual $\boldsymbol{r}$ in surge ($X$), sway ($Y$), and yaw ($N$) DoFs, shown along with GP confidence intervals (shaded areas) and the true fault-induced disturbance $\boldsymbol{d}_f$.

### B. Teleoperation Experiments

The performance of the proposed framework was first validated under open-loop control of the vehicle horizontal motion. In particular, the vehicle was teleoperated by applying surge, sway, and yaw commands through a joystick. Two fault scenarios were investigated, namely complete loss of effectiveness in thruster 2 ($\lambda_2 = 0$) and thruster 3 ($\lambda_3 = 0$).

In the first scenario, the fault was introduced at $t = 20$ s while a negative surge command was applied. The fault was successfully detected and isolated using the GLRT defined in Eq. 18. Specifically, the test statistic associated with the affected actuator, $T_2$, exceeded the detection threshold $\delta$ approximately 2 s after fault injection, as illustrated in Fig. 5. Following detection, the MLE of the corresponding thruster effectiveness converged to the true value (Fig. 5), demonstrating the capability of the proposed approach to accurately estimate the fault magnitude. Fault detection and estimation are enabled by the residual $\boldsymbol{r}$, constructed from the MHE-based lumped disturbance estimate and the GP posterior distribution of the unmodeled dynamics. As shown in Fig. 5, the residual effectively captured the true fault-induced disturbance $\boldsymbol{d}_f$. It should be noted that, due to similarities among the columns of the thruster allocation matrix, the GLRT statistics associated with healthy thrusters also increased after fault injection. Nevertheless, the statistic $T_2$ remained distinctly dominant, thereby enabling reliable fault isolation under the single-fault assumption.

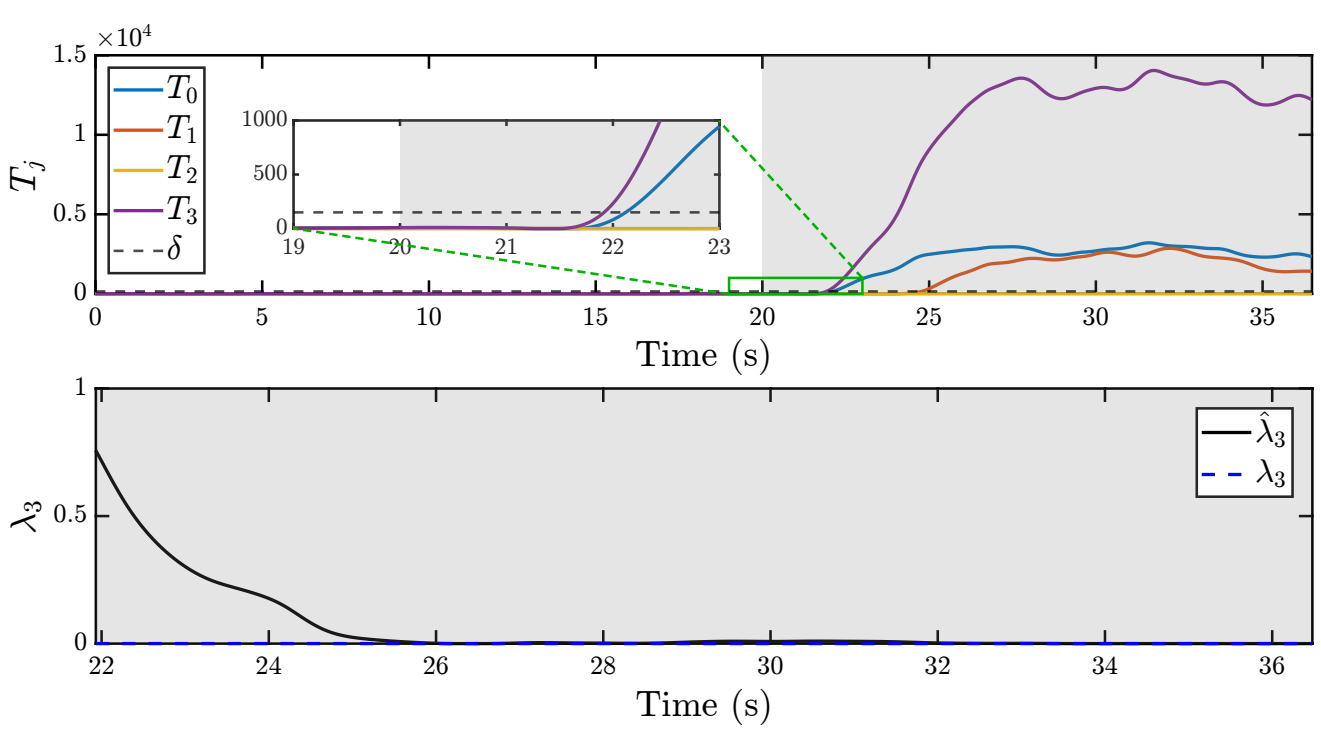

Fig. 6. Second teleoperation experiment with complete failure of Thruster 3. From top to bottom: GLRT statistics $T_j$ of the horizontal thrusters, with the shaded region indicating operation after fault injection; and the MLE estimate $\hat{\lambda}_3$ together with the true value after fault detection.

Similar behavior was observed in the second scenario involving complete failure of thruster 3. After fault introduction, the vehicle was commanded to move in the positive surge direction. As depicted in Fig. 6, the fault was rapidly detected and isolated, and the effectiveness of thruster 3 was accurately estimated.

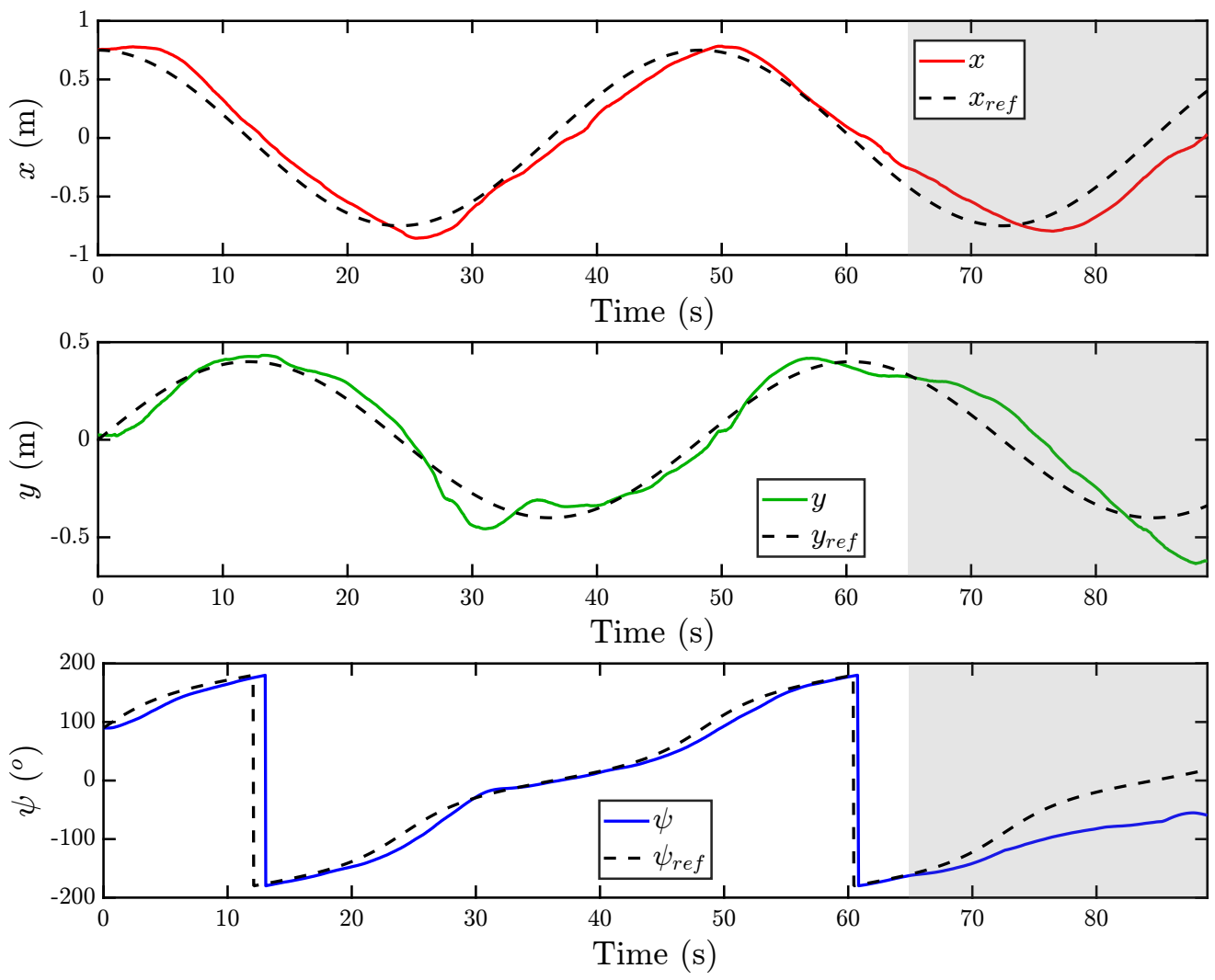

Fig. 7. Actual and reference trajectories, with the shaded region denoting operation after fault injection ($\lambda_0 = 0.25$).

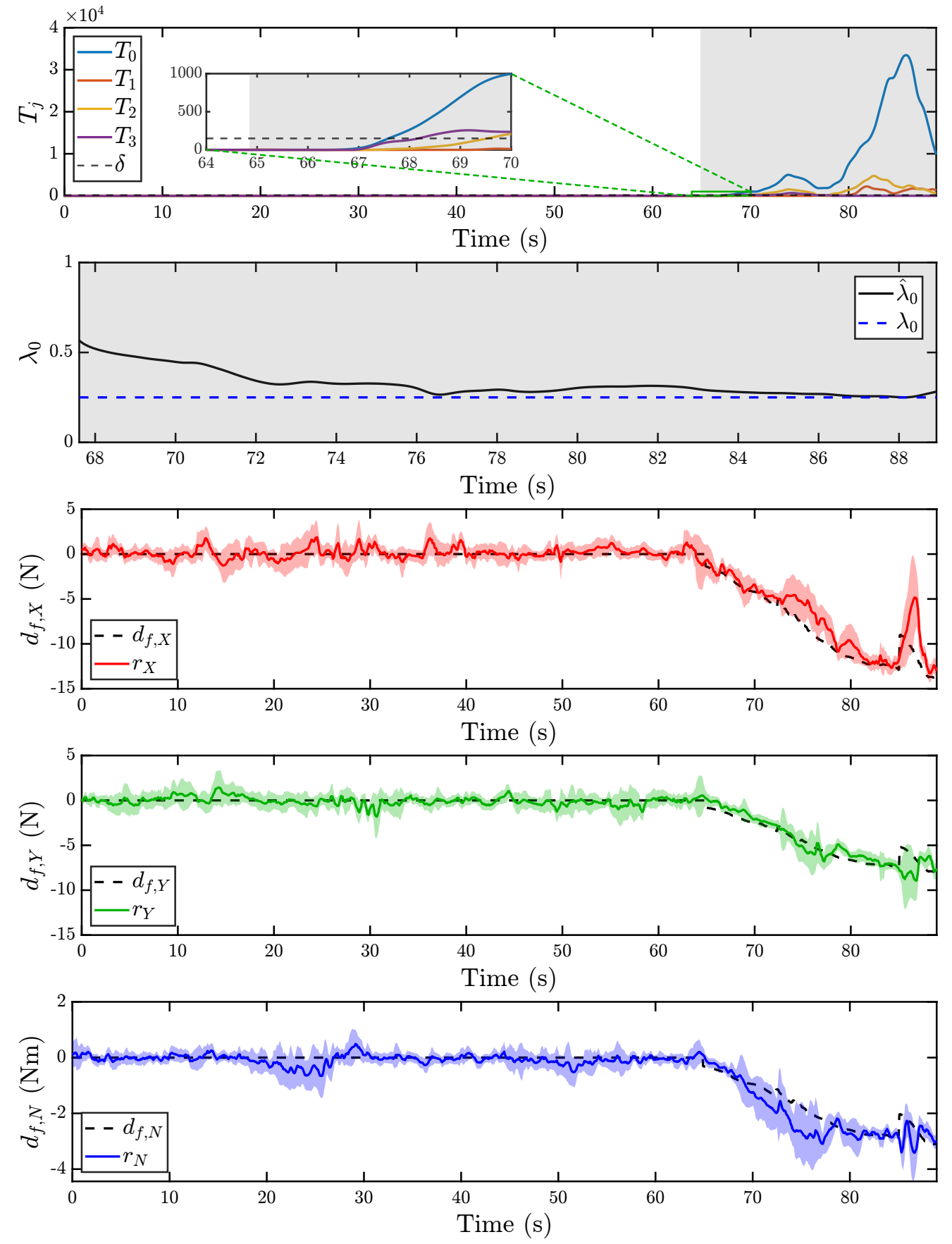

Fig. 8. Closed-loop experiment with failure of Thruster 0 ($\lambda_0 = 0.25$).

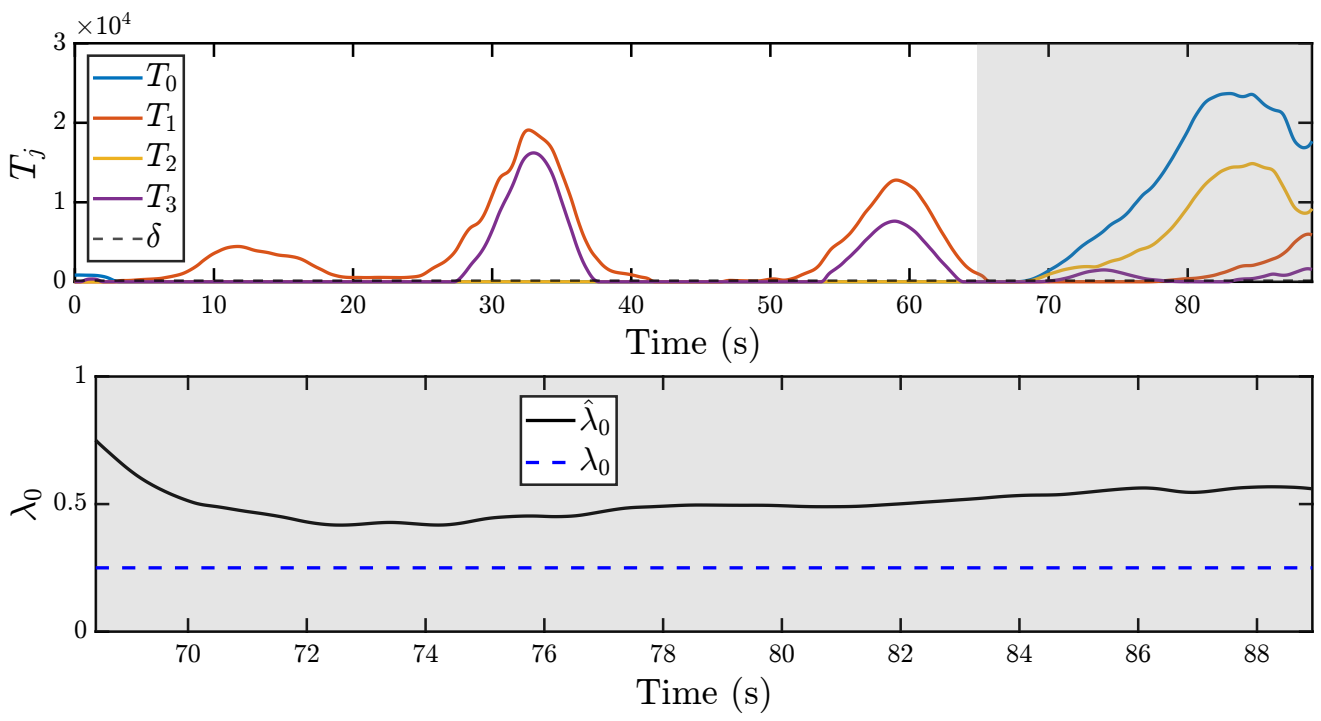

Fig. 9. Closed-loop experiment with failure of Thruster 0 ($\lambda_0 = 0.25$) without considering the unmodeled dynamics.

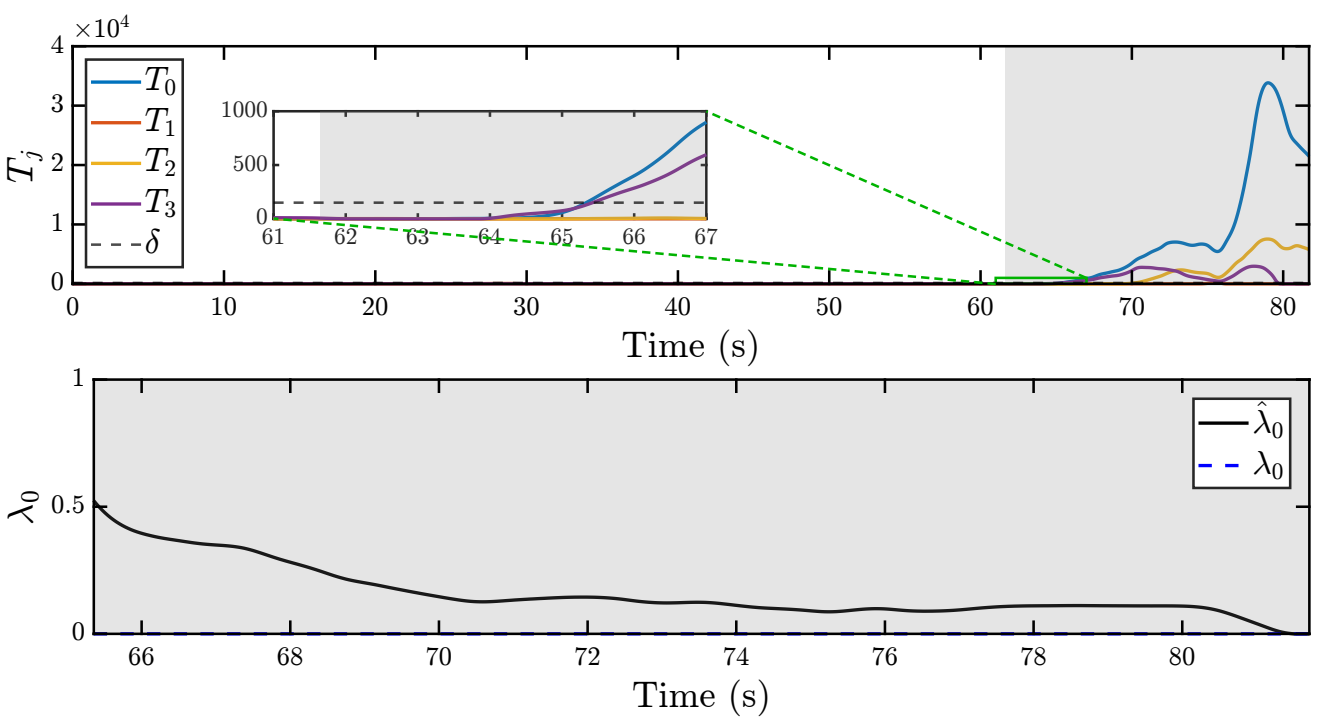

Fig. 10. Closed-loop experiment with complete failure of Thruster 0 ($\lambda_0 = 0$).

### C. Closed-Loop Experiments

The proposed framework was next assessed under closed-loop control, where the vehicle was commanded to track an ellipsoidal reference trajectory in the horizontal plane, defined by $x_r = 0.75\cos(0.13t)$ and $y_r = 0.4\sin(0.13t)$. The reference yaw angle was aligned with the trajectory tangent, i.e., $\psi_r = \text{atan2}(\dot{y}_r, \dot{x}_r)$. Under normal conditions, satisfactory trajectory-tracking performance was achieved using a PID controller, as illustrated in Fig. 7, until approximately $t = 65$ s, when the effectiveness of thruster 0 was reduced to $\lambda_0 = 0.25$. The GLRT framework rapidly identified the fault, while the corresponding MLE converged to the true fault magnitude, as shown in Fig. 8. Consistent with the teleoperation experiments, the residual $\boldsymbol{r}$ reliably approximated the temporal evolution of the fault-induced disturbance $\boldsymbol{d}_f$ (Fig. 8), thus enabling successful fault detection and quantification during closed-loop trajectory tracking.

To demonstrate the necessity of learning the unmodeled dynamics, a comparative analysis was performed based on data recorded during the aforementioned closed-loop experiment. In this setting, the unmodeled dynamics were assumed negligible, and the residual was therefore constructed solely from the MHE lumped disturbance estimates, i.e., $\boldsymbol{r}_k = \hat{\boldsymbol{d}}_k$. In contrast to the proposed GP-based approach, the residual standard deviation was constant, with values equal to 2.0 N, 2.0 N, and 0.5 Nm in the surge, sway, and yaw DoFs, respectively. The corresponding GLRT statistics are presented in Fig. 9. Without accounting for the unmodeled dynamics, several statistics associated with different thrusters exceeded the threshold $\delta$ prior to fault injection, thereby producing false alarms. After the failure of thruster 0, the statistic $T_0$ correctly crossed the threshold; however, the corresponding MLE of the fault magnitude remained inaccurate without compensation for the unmodeled dynamics (Fig. 9). It is highlighted that reducing the false-alarm rate by tuning the threshold $\delta$ is inherently challenging, as the magnitudes of $T_j$ associated with false alarms and true faults are comparable.

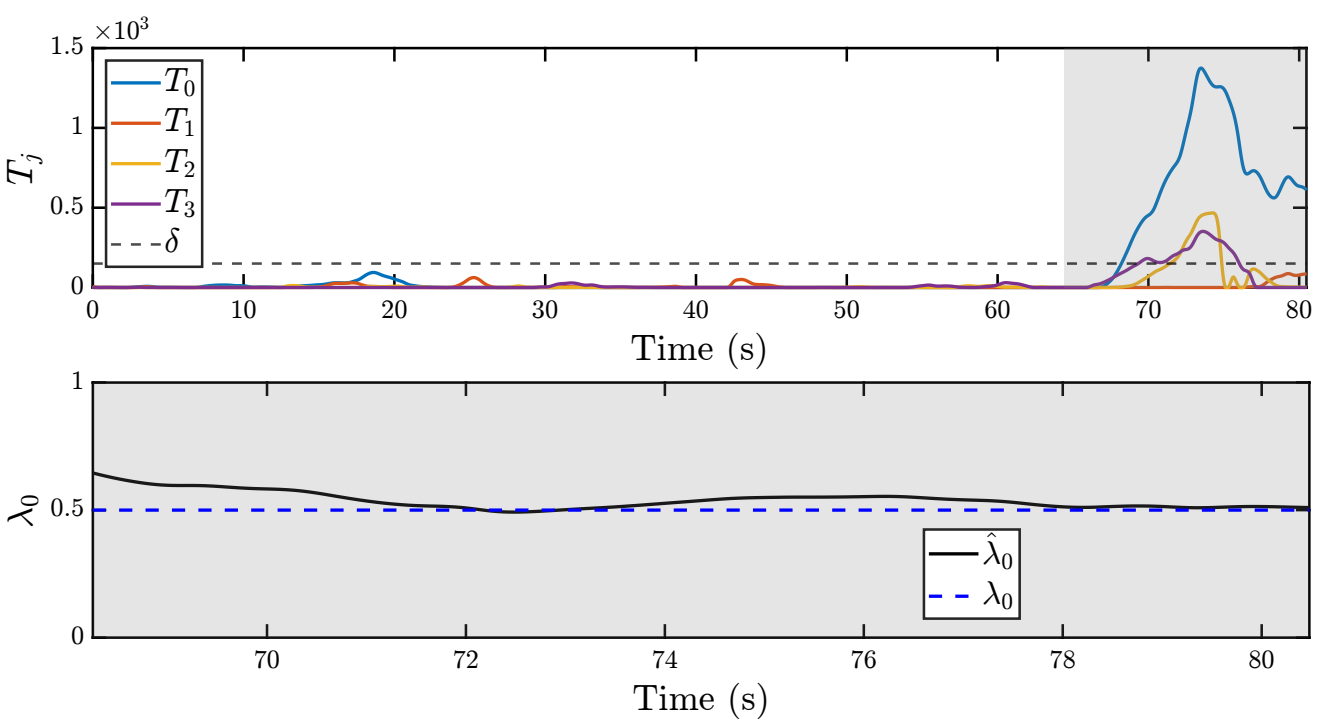

Fig. 11. Closed-loop experiment with failure of Thruster 0 ($\lambda_0 = 0.5$).

Finally, the robustness of the proposed method was further evaluated through two additional fault scenarios while the vehicle was commanded to track the same ellipsoidal reference trajectory. Specifically, complete failure and partial loss of effectiveness were introduced in thruster 0, corresponding to $\lambda_0 = 0$ and $\lambda_0 = 0.5$, respectively. As illustrated in Fig. 10 and Fig. 11, the proposed framework successfully identified the injected faults in both cases.

## VI. CONCLUSIONS

In this paper, we presented a model-based FDD framework for UVs that enables reliable fault detection, isolation, and quantitative fault severity estimation in the presence of unmodeled dynamics. By integrating MHE-based disturbance estimation, GP regression for the approximation of the unmodeled dynamics, and GLRT-based statistical hypothesis testing, the proposed approach achieves robust diagnostic performance, as validated across different scenarios.

Regarding our future work, we aim to address the simultaneous occurrence of multiple faults. In addition, we will extend the methodology to account for considerable environmental disturbances through online training of the GPs with a safe update mechanism. Finally, we intend to integrate the proposed FDD framework with an active fault-tolerant control scheme to enhance control performance.

## ACKNOWLEDGMENT

This work is supported in part by the NYUAD Center for Artificial Intelligence and Robotics, funded by Tamkeen under the Research Institute Award CG010.